\documentclass[preprint]{article}

\PassOptionsToPackage{numbers, compress}{natbib}

\usepackage{neurips_2025}

\usepackage{wrapfig}
\usepackage[utf8]{inputenc} 
\usepackage[T1]{fontenc}    
\usepackage{hyperref}       
\usepackage{url}            
\usepackage{booktabs}       
\usepackage{amsfonts}       
\usepackage{nicefrac}       
\usepackage{microtype}      
\usepackage{xcolor}         
\usepackage{amssymb}
\usepackage{breqn}
\usepackage{mathtools}
\usepackage{subcaption}
\usepackage{hyperref}
\usepackage{algorithm}
\usepackage{algpseudocode}
\usepackage{enumitem}
\usepackage{pifont}
\usepackage[absolute,overlay]{textpos}
\usepackage{soul}
\usepackage{booktabs}  
\usepackage{array}     
\usepackage{animate}
\usepackage{multirow}
\usepackage{booktabs} 
\usepackage{bbm}

\input{mysymbol.sty}

\title{Graph Guided Diffusion: Unified Guidance for Conditional Graph Generation}

\makeatletter
\def\thanks#1{\protected@xdef\@thanks{\@thanks
        \protect\footnotetext{#1}}}

\author{%
  Victor M. Tenorio\\
  King Juan Carlos University\\
  \texttt{victor.tenorio@urjc.es}\\
  \And
  Nicolas Zilberstein\\
  Rice University\\
  \texttt{nzilberstein@rice.edu}\\
  \AND
  Santiago Segarra \\
  Rice University\\
  \texttt{segarra@rice.edu} \\
  \And
  Antonio G. Marques \\
  King Juan Carlos University\\
  \texttt{antonio.garcia.marques@urjc.es}\\
}

\begin{document}

\maketitle

\begin{abstract}
Diffusion models have emerged as powerful generative models for graph generation, yet their use for conditional graph generation remains a fundamental challenge. In particular, guiding diffusion models on graphs under arbitrary reward signals is difficult: gradient-based methods, while powerful, are often unsuitable due to the discrete and combinatorial nature of graphs, and non-differentiable rewards further complicate gradient-based guidance.
We propose Graph Guided Diffusion (GGDiff), a novel guidance framework that interprets conditional diffusion on graphs as a stochastic control problem to address this challenge. GGDiff unifies multiple guidance strategies, including gradient-based guidance (for differentiable rewards), control-based guidance (using control signals from forward reward evaluations), and zero-order approximations (bridging gradient-based and gradient-free optimization).
This comprehensive, plug-and-play framework enables zero-shot guidance of pre-trained diffusion models under both differentiable and non-differentiable reward functions, adapting well-established guidance techniques to graph generation — a direction largely unexplored. Our formulation balances computational efficiency, reward alignment, and sample quality, enabling practical conditional generation across diverse reward types. We demonstrate the efficacy of GGDiff in various tasks, including constraints on graph motifs, fairness, and link prediction, achieving superior alignment with target rewards while maintaining diversity and fidelity.
\end{abstract}

\section{Introduction}

Diffusion models have recently shown great promise for graph generation, enabling the synthesis of realistic graph structures across diverse domains such as drug design~\cite{yang2024molecule}, social networks~\cite{grover2019graphite}, and molecular dynamics~\cite{hoogeboom2022equivariant}. 
A key motivation behind these models is their ability to serve as \emph{flexible generative priors}, capturing complex dependencies in both graph topology and node features. 
However, most existing graph diffusion models focus on unconditional or controllable generation under simple objectives. Incorporating more \emph{general forms of rewards or constraints}, such as enforcing specific structural properties, functional motifs, or domain-specific validity criteria like fairness, remains an essential and open challenge.

Recent advances in conditional graph generation typically modify the diffusion trajectory using a conditional gradient to steer the process toward sampling from the desired conditional distribution. DiGress~\cite{vignac2023digress} combines a learnable regressor with classifier guidance~\cite{ho2021classifier}, while LGD~\cite{zhou2024unifying} adopts a similar gradient-based strategy in a latent space instead of a discrete domain like DiGress.
However, these approaches require differentiable constraints, which limits their applicability in more complex graph generation tasks where constraints might be black-box functions without tractable gradients or involve discrete structures.
Closer to our approach, PRODIGY~\cite{sharma2024diffuse} enforces hard constraints through projected sampling along the diffusion trajectory using the bisection method~\cite{boyd2004convex}. 
This requires closed-form projection operators for a time-efficient implementation, which are often unavailable for complex constraints, necessitating expensive solvers, and limiting the applicability to more sophisticated conditions.
As a result, no existing method can flexibly and effectively handle arbitrary, non-differentiable, or complex constraints in the sampling process for conditional graph generation.

In this work, we propose Graph Guided Diffusion (GGDiff), a general guidance framework for graph generation that interprets conditional graph generation as a stochastic optimal control (SOC) problem. 
By casting the task as a control problem, we reformulate guided diffusion as a conditional generation process with an additional control variable defined as a linear drift term. 
Inspired by recent advances in SOC for diffusion models~\cite{pandey2025variational,huang2024symbolic,routrb}, we optimize this control via path integral control, which provides an analytical yet intractable gradient. 
To overcome this limitation, we introduce gradient-free approximations based on zeroth-order (ZO) optimization techniques~\cite{liu2020primer, liu2018zeroth}.
Our formulation generalizes several gradient-free strategies introduced previously~\cite{huang2024symbolic} and offers new possibilities. 

GGDiff unifies various existing guidance methods in a single framework and is plug-and-play, allowing zero-shot guidance of pre-trained diffusion models under both differentiable and non-differentiable reward functions. 
We validate the advantages of GGDiff through extensive experiments on a wide range of constraints, including structural constraints, fairness, and link prediction.
Our results demonstrate GGDiff's versatility in guiding graph generation not only towards constraints previously explored in the literature (beating current state-of-the-art architectures) but also towards arbitrary, user-defined desired outcomes (such as fair or incomplete graphs), effectively balancing precise outcome satisfaction with the preservation of the underlying graph family's characteristics.

To summarize, our contributions are threefold:
\vspace{-0.2cm}
\begin{itemize}
    \setlength\itemsep{-0.1em}
    \item We propose GGDiff, a framework for conditional graph generation that handles both differentiable and non-differentiable rewards by reformulating the problem as a stochastic optimal control (SOC) task.
    \item We introduce a general gradient-free ZO optimization formulation to handle non-differentiable rewards, enabling optimization without requiring tractable gradients.
    \item We conduct extensive experiments on structural, fairness, and link prediction constraints, demonstrating GGDiff's superior performance and flexibility over existing methods.
\end{itemize}

\begin{figure}
    \centering
    \includegraphics[width=0.65\linewidth]{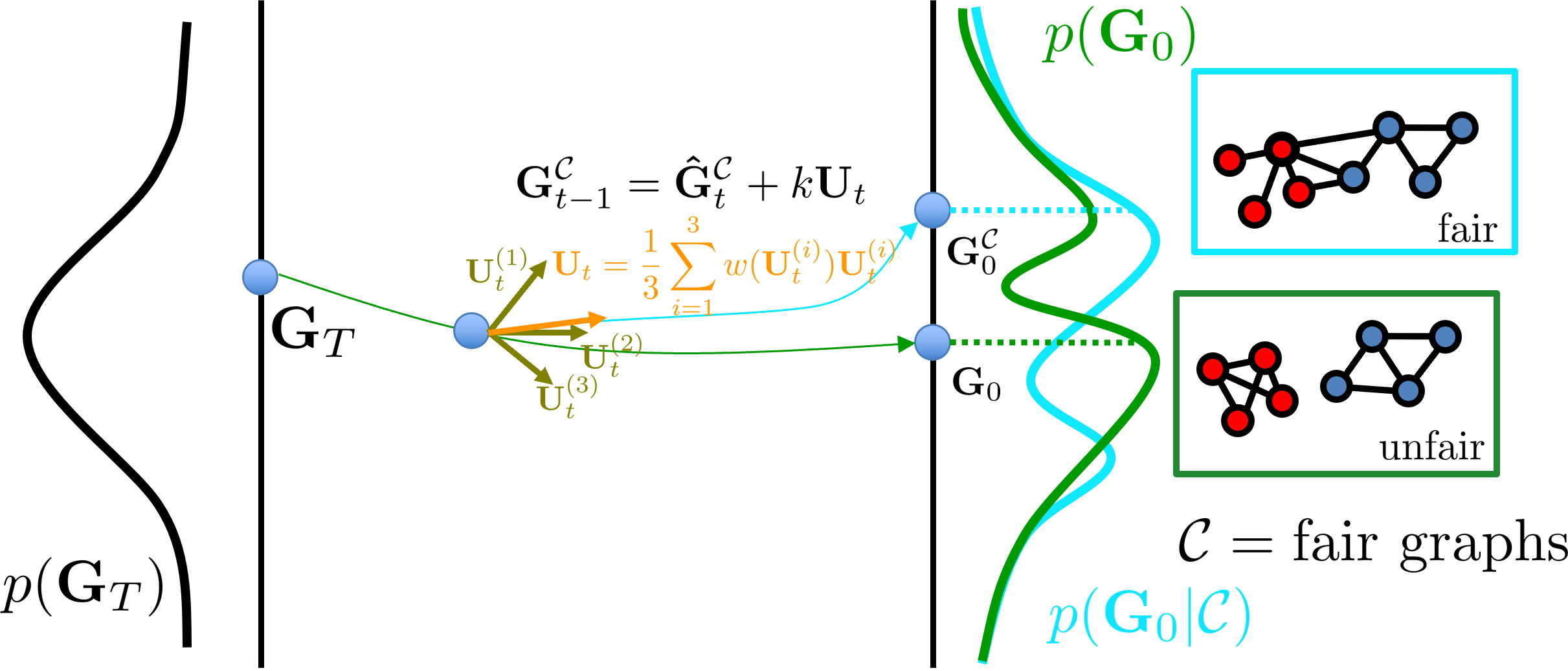}
    \caption{\small{Illustration of GGDiff, a method that guides the generation of graphs to satisfy a set of constraints (in this case, the constraint is fairness). 
    The guidance $\bbU_t$ is a local direction obtained via SOC, and approximated using ZO techniques, like the multi-point estimate shown here.}}
    \label{fig:teaser}
    \vspace{-4mm}
\end{figure}




\section{Background and Related Works}

We review graph diffusion models in the continuous domain in Section~\ref{subsec:background_diffusion}, and then explain how they can be used in the context of inverse problems in Section~\ref{subsec:background_posterior}

\subsection{Diffusion Models on Graphs} 
\label{subsec:background_diffusion}

Diffusion models~\citep{sohl2015deep, ho2020denoising, song2020score} are composed of two processes:
i) a forward process that starts with clean data and gradually adds noise; and
ii) a reverse process that learns to generate new data by iteratively denoising its diffused version. 
While graph diffusion models have been developed in both \emph{continuous}~\cite{niu2020permutation, jo2022score} and \emph{discrete} domains~\cite{vignac2023digress, chen2023efficient}, in this work we focus on the continuous case.
We represent a graph as $\bbG_0 = \{\bbX_0, \bbA_0\}$, where $\bbX_0 \in \reals^{N\times F}$ are node features and $\bbA_0 \in \reals^{N\times N}$ is the weighted adjacency matrix.
Then, following GDSS~\citep{jo2022score}, the forward diffusion process is defined by the stochastic differential equation
$\text{d}\bbG_t = -\frac{1}{2}\beta(t)\bbG_t\,\mathrm{d}t + \sqrt{\beta(t)}\,\mathrm{d}\bbW_t, \quad t \in [0, T]$,
where $\beta(t)$ controls the noise schedule and is given by
$\beta(t) := \beta_{\mathrm{min}} + (\beta_{\mathrm{max}} - \beta_{\mathrm{min}})\frac{t}{T}$.
Here, $\bbW_t$ denotes standard Brownian motion. This process is designed such that the distribution of $\bbG_T$ converges to a standard Gaussian as $t \to T$.
Notice that by construction, the forward process allows for interaction between the adjacency matrix and the node features.
Based on this forward process, we define the reverse process as $\text{d}\bbG_t = [-\frac{1}{2}\bbG_t -g(t)^2\nabla_{\bbG_t}\log p(\bbG_t)]\text{d}t + g(t)\text{d}\bbW_t$, where $\nabla_{\bbG_t}\log p(\bbG_t)$ is the \emph{score function}, which is unknown, and $g(t) = \sqrt{\beta(t)}$.
In particular, GDSS considers two different score functions, namely $\nabla_{\bbA_t}\log p(\bbA_t)$ and $\nabla_{\bbX_t}\log p(\bbX_t)$.

Notice that the requirement for sampling, i.e., running the reverse process, is to have access to the score functions $\nabla_{\bbA_t}\log p(\bbA_t)$ and $\nabla_{\bbX_t}\log p(\bbX_t)$, which are generally unknown.
Instead, we approximate them with \emph{score networks} $\bbepsilon_{\bbtheta_A}(\bbA_t, t) \approx -\sigma_t \nabla_{\bbA_t}\log p(\bbA_t)$ and $\bbepsilon_{\bbtheta_X}(\bbX_t, t) \approx -\sigma_t \nabla_{\bbX_t}\log p(\bbX_t)$, and learn them by minimizing the denoising score-matching loss~\citep{vincent2011connection}.
After training, samples are generated using samplers like DDPM~\citep{ho2020denoising} and DDIM~\citep{song2020denoising}.

\subsection{Controllable Generation of Graphs With Continuous Diffusion Models}
\label{subsec:background_posterior}

Given a condition $\ccalC$ and a reward function $r(\bbG_0)$ that quantifies how close the sample $\bbG_0$ is to meeting $\ccalC$, our objective is to generate graphs $G_0$ that maximize the reward function.
From a Bayesian perspective, this problem boils down to sampling from the posterior $p(\bbG_0|\ccalC) \propto p(\ccalC|\bbG_0)p(\bbG_0)$ where $p(\ccalC|\bbG_0) \propto \exp{(r(\bbG_0))}$ is a likelihood term and $p(\bbG_0)$ is a prior given by the pre-trained diffusion model.
We now describe previous works for both differentiable and non-differentiable reward functions.

\paragraph{Controllable generation with differentiable rewards.}
A plethora of works have proposed solutions to Problem 1 in the context of inverse problems in images~\citep{chung2022diffusion, mardani2023variational, zilberstein2024solving, zilberstein2024repulsive}.
In general, these works assume a differentiable reward -- the likelihood associated with a noisy measurement--, which allows us to compute the \textit{conditional score} at noise level $t$ obtained via Bayes’ rule 
\begin{equation}\label{eq:score_post}
    \nabla_{\bbG_t}\log p(\bbG_t|\ccalC) =  \nabla_{\bbG_t}p(\ccalC|\bbG_t) +  \nabla_{\bbG_t}\log p(\bbG_t).
\end{equation}
This naturally allows us to incorporate the diffusion model as the prior. 
However, the score associated to the likelihood term is intractable, as seen from $p(\ccalC|\bbG_t) = \int p(\ccalC|\bbG_0)p(\bbG_0|\bbG_t)\text{d}\bbG_0$.
To circumvent this, different works proposed to use Gaussian approximation of $p(\bbG_0|\bbG_t)$ centered at the MMSE denoiser, which can be computed using Tweedie's formula $\mathbb{E}[\bbG_0|\bbG_t] = \frac{1}{\alpha_t}\left(\bbG_t + \sigma_t^2 \nabla_{\bbG_t} \log p (\bbG_t, t)\right)$.
In the context of graph, DiGress~\cite{vignac2023digress} 
and LGD~\cite{zhou2024unifying} incorporates a guidance via a learned model (similar to classifier-free guidance), requiring an extra learnable model.
However, this approximation remains largely unexplored in the context of graph inverse problems, mainly because it requires (and assumes) \emph{differentiable} rewards, which in general are not available for graph generation~\citep{you2018graph}.

\paragraph{Controllable generation with non-differentiable rewards.} 
Recently, PRODIGY~\cite{sharma2024diffuse} explored an alternative for controllable graph generation with non-differentiable constraints.
Their core idea involves an unconditional generation step from $\bbG_t$ to produce a candidate $\hat{\bbG}_{t-1}$, followed by a projection step. 
Formally, after obtaining $\hat{\bbG}_{t-1}$, they apply a projection operator $\Pi_{\ccalC}(\hat{\bbG}_{t-1}) = \argmin_{\bbZ \in \ccalC}||\bbZ - \hat{\mathbf{G}}_{t-1}||_2^2$. While direct projection ensures the sample satisfies $\ccalC$, applying it fully at each noise level can disrupt the learned reverse trajectory's smoothness. 
To address this, PRODIGY proposes a partial update: $\mathbf{G}_{t-1} \leftarrow\left(1-\gamma_t\right) \hbG_{t-1}+\gamma_t \Pi_{\mathcal{C}}\left(\hbG_{t-1}\right)$, where $\gamma_t$ balances constraint adherence with the original diffusion path.
Although PRODIGY implements this constrained step efficiently for simple constraints with closed-form expression for $\Pi_{\ccalC}(.)$, its reliance on such operators limits applicability to general reward functions without incurring prohibitive runtime.
Furthermore, applying the projection operator directly to the noisy variable $\bbG_t$ rather than the denoised estimate $\mathbb{E}[\bbG_0|\bbG_t]$ is misaligned with the reward domain, which is defined at the data level ($t=0$).
In addition, the work in~\cite{madeira2024generative} considered more complex constraints.
In a nutshell, they combine projection operators combined with an edge-absorbing model.
While effective, this method is computationally demanding due to its combinatorial nature and using discrete diffusion models.

\section{Controllable Generation of Graphs With General Rewards}

In Section~\ref{subsec:cond_gen_soc}, we formulate the generation of graph conditionals as a SOC problem.
Then, in Section~\ref{subsec:greedy_sol} we propose different approximate solutions to design the control for conditional graph generation: first, in Section~\ref{subsubsec:differentiable} we introduce our approximation for differentiable rewards; second, in Section~\ref{subsubsec:zero_order} we propose a ZO approximation, which unifies several existing guidance policies for non-differentiable rewards.

\subsection{Conditional Generation: A SOC Approach}
\label{subsec:cond_gen_soc}

The goal of our method is to \emph{steer} a pre-trained diffusion model to sample from the posterior distribution.
Importantly, we seek an algorithm that can handle general reward functions, even non-differentiable ones.
To tackle this, we proposed to leverage SOC~\cite{van2007stochastic}.
In particular, given an uncontrolled diffusion process $\ccalQ$ (defined in Section~\ref{subsec:background_diffusion}), we define a controlled one $\ccalQ^\ccalC$ given by
%
\begin{align}
\label{eq:q_u}
    \ccalQ^\ccalC: \text{d}\bbG_t^\ccalC = \left[-\frac{1}{2}\bbG_t^\ccalC -g(t)^2\nabla_{\bbG_t^\ccalC}\log p(\bbG_t^\ccalC) + g(t) \bbU(\bbG_t^\ccalC, t)\right]\text{d}t + g(t)\text{d}\bbW_t, \quad t \in [T, 0].
\end{align}
%
Thus, the goal is to design the control $\{\bbU(\bbG_t^\ccalC, t)\}_{t\in[0,T]}$ to modify the trajectory of the controlled process $\ccalQ^{\ccalC}$ such that the generated samples belong to the target distribution. 
We formalize this as a SOC problem, where we solve the following optimization problem
\begin{equation}
\label{eq:SOC}
\min _{\bbU \in \mathcal{U}} \mathbb{E}\left[\int_0^T \lambda\frac{||
\bbU\left(\bbG_t^\ccalC, t\right)||^2_F}{2} \mathrm{d} t - r\left(\bbG_0^\ccalC \right)\right]\quad \text{s.t.}\; \ccalQ^\ccalC.
\end{equation}
The terminal cost in~\eqref{eq:SOC} represents a desired constraint for the final state $\bbG_0$ quantified by the reward $r(.)$, which is maximized (thus, the negative sign), while the transient term is a regularization term that penalizes large deviation from the uncontrolled process by promoting the energy of the controller in~\eqref{eq:q_u} to be small.
The solution of~\eqref{eq:SOC} is given by the Feynman-Kac formula, a well-known result from the optimal control theory~\cite{pavon1989stochastic}, given by
\begin{equation}
\label{eq:optimal_control}
\bbU^*(\bbG_t^\ccalC , t) = -g(t)\nabla_{\bbG_t^\ccalC } \log \mathbb{E}_{p^{\mathrm{pre}}} \left[ \exp\left( \frac{-r(\bbG_0^\ccalC )}{\lambda} \right) \,\Big|\, \bbG_t^\ccalC \right].
\end{equation}
The solution in~\eqref{eq:optimal_control} is obtained as the solution of the \emph{linear} version of the Hamilton-Jacobi-Bellman (HJB) equation~\cite{evans2022partial}, obtained after the exponential transformation; we deferred to Appendix~\ref{app:oc_theory} for more details on the derivation. 

Given the optimal control, we now focus on how to implement it.

\subsection{Estimation of the Optimal Control: A Greedy Solution}
\label{subsec:greedy_sol}

Although the expression for the optimal control derived from the Feynman-Kac formula~\eqref{eq:optimal_control} is theoretically exact, its direct computation is often intractable. 
Evaluating the expectation and its gradient would require simulating numerous trajectories of the uncontrolled process from the current state $\bbG_t^{\ccalC}$ to the final state $\bbG_0^{\ccalC}$ at each step of the generation process to estimate $p^{\mathrm{pre}}$, and then backpropagating through the diffusion trajectory.
This is computationally prohibitive.

We resort to a \emph{greedy} approximation strategy to overcome this. 
This approach simplifies the problem by approximating the complex gradient of the log-expectation term in~\eqref{eq:optimal_control} using primarily the current state information $\bbG_t^{\ccalC}$ and a one-step estimate of the clean sample $\hat{\bbG}_0^{\ccalC}$. 
Such an approximation implies that the control decision at time $t$ does not fully account for the entire future trajectory, potentially leading to suboptimal choices, especially in the early stages of the reverse diffusion process. 
However, the impact of such approximation errors may often diminish as $t \to 0$ and the state $\bbG_t^{\ccalC}$ gets closer to the data. 
We now detail this approximation for the cases of $(i)$ differentiable rewards and $(ii)$ non-differentiable counterparts.

\subsubsection{Differentiable Rewards}
\label{subsubsec:differentiable}

When the reward function $r(\cdot)$ is differentiable, we can derive a tractable approximation for the optimal control $\bbU^*(\bbG_t^\ccalC , t)$.
The primary challenge lies in evaluating the gradient of the log-expectation term. 
To circumvent this, we use Tweedie's formula (see Section~\ref{subsec:background_diffusion}) to compute the MMSE denoiser $\mathbb{E}[\bbG_0^\ccalC |\bbG_t^\ccalC ] = \hat{\bbG}_0^\ccalC (\bbG_t^\ccalC )$ and approximate the conditional expectation in~\eqref{eq:optimal_control} as
\begin{equation}
 \mathbb{E}_{p^{\mathrm{pre}}} \left[ \exp\left( \frac{-r(\bbG_0^\ccalC )}{\lambda} \right) \,\Big|\, \hbG_t^\ccalC \right] \approx \exp\left( \frac{-r(\hat{\bbG}_0^\ccalC (\hbG_t^\ccalC ))}{\lambda} \right),
 \label{eq:direct_expectation_approximation}
\end{equation}
where the underlying assumption is that $p(\bbG_0^{\ccalC}|\hbG_t^{\ccalC}) = \delta(\bbG_0^{\ccalC} - \hat{\bbG}_0^{\ccalC}(\hbG_t^{\ccalC}))$ with $\delta(.)$ denoting a Dirac delta function.
This approximation becomes increasingly better as $t \to 0$ (i.e., towards the end of the reverse diffusion process), as $\hat{\bbG}_0^{\ccalC}(\hbG_t^{\ccalC})$ becomes a better estimate of $\bbG_0^{\ccalC}$.

Substituting this approximation into the exact optimal control formula in~\eqref{eq:optimal_control} leads to
\begin{align}
    \bbU^*(\hbG_t^{\ccalC}, t) &\approx \frac{g(t)}{\lambda}\nabla_{\hbG_t^{\ccalC}} r(\hat{\bbG}_0^{\ccalC}(\hbG_t^{\ccalC})).
    \label{eq:greedy_approx_control_final_direct}
\end{align}
This final expression provides a tractable, greedy approximation for the optimal control. 
The control term now directly involves the gradient of the reward function $r(\cdot)$ evaluated at the one-step denoised estimate $\hat{\bbG}_0^{\ccalC}$.  The term $1/\lambda$ acts as a scaling factor for the guidance.
This formulation resembles guidance techniques in diffusion models, as observed by~\cite{huang2024symbolic, uehara2025inference}.
For example, if the reward $r(\bbG_0)$ is proportional to the log-likelihood of a condition $\mathcal{C}$, that is, $r(\bbG_0) \propto -\log p(\mathcal{C}|\bbG_0)$, then the optimal controls boils down to the DPS approximation~\cite{chung2022diffusion}.

\subsubsection{Non-differentiable Rewards}
\label{subsubsec:zero_order}

In many practical scenarios of controlled graph generation, the reward function $r(\cdot)$ is non-differentiable with respect to the generated graph $\bbG_0^{\ccalC}$, 
rendering gradient-based approximations like~\eqref{eq:greedy_approx_control_final_direct} intractable.

To address this, we propose to determine the control input $\bbU(\bbG_t, t)$ using an approach inspired by gradient-free optimization methods~\cite{larson2019derivative} and ZO optimization~\cite{liu2020primer}.
The objective at each time $t$ is to find a control $\bbU(\bbG_t, t)$ that steers the diffusion trajectory towards graphs yielding a high reward $r(\bbG_0^{\ccalC})$.
Similar to the differentiable case, we use Tweedie's formula to compute a one-step denoised version of the final graph to evaluate the reward at each time step.
Given this approximation, we formally seek to find a direction $\bbU^*_t$
\begin{equation}
    \bbU_{t}^* = \argmax_{\bbU_t} \, r\left(\hat{\bbG}_0^{\ccalC}(\hbG_{t}^{\ccalC} +  \mu\bbU_t)\right),
    \label{eq:zo_objective}
\end{equation}
Here, $\hbG_{t}^{\ccalC} +  \mu\bbU_t$ denotes the perturbed version of $\hbG_t^{\ccalC}$, which is the generated graph with the reference model at time $t$ (before applying the guidance) following the control direction $\bbU_t$, and $\mu$ is a smoothing parameter (which depends on the noise schedule of the diffusion process).
To find $\bbU^*_t$, we define a general ZO estimator for the gradient of the reward that depends on evaluations of $r(.)$ as
\begin{equation}
    \label{eq:zo_general_form}
    \hat{\nabla} r(\hbG_t^\ccalC) := \mathbb{E}_{\bbU_t \sim \mathcal{D}} \left[ w(\bbU_t) \, r\left(  \hat{\bbG}_0^{\ccalC}(\hbG_{t}^{\ccalC} +  \mu\bbU_t) \right) \cdot \bbU_t \right],
\end{equation}
where \( \mathcal{D} \) is a distribution over directions (typically Gaussian) and \( w(\bbU_t) \) is a direction-dependent weighting function. 
Notably, this formulation unifies several previous gradient-free estimators.
However, it is important to remark that traditional ZO optimization assumes the objective is differentiable but the gradient is inaccessible.
In contrast, in our setting the reward function $r(.)$ is inherently non-differentiable, often defined via a discrete or combinatorial metric over generated graphs. 
Nevertheless, we treat the reward as a black-box function and employ randomized directional evaluations to define a \emph{pseudo-gradient} direction that can guide the controlled process.
Thus, the ZO estimator in~\eqref{eq:zo_general_form} should be interpreted as a surrogate direction that correlates with improvements in the reward, rather than an unbiased estimator of a true gradient. 

We now present three practical ZO estimators that instantiate~\eqref{eq:zo_general_form}.

\paragraph{One-point (and two-point) gradient estimators.}
The one-point estimator samples a single perturbation direction \( \bbU_t \sim \mathcal{N}(\bbzero, \mathbf{I}) \) and evaluates the reward by perturbing the unconditional generated graph with this single direction. 
The estimated gradient is given by
\begin{equation}
    \label{eq:1point_estimate}
    \hat{\nabla} r(\hbG_t^\ccalC) = \frac{\phi(d)}{\mu} \, r\left( \hat{\bbG}_0^{\ccalC}(\hbG_{t}^{\ccalC} +  \mu\bbU_t) \right) \cdot \bbU_t,
\end{equation}
where \( \phi(d) \) is a scaling factor that depends on $\ccalD$; for $\ccalD$ Gaussian, we have $\phi(d) = 1$.
This control corresponds to $w(\bbU_t) = \frac{\phi(d)}{\mu}$.
In classical ZO, this estimator is an unbiased estimator of the smoothed version of $r(.)$ over a random perturbation, i.e., $\mathbb{E}_{\bbU_t\sim \ccalD}[r(\hat{\bbG}_0^{\ccalC}(\hbG_{t}^{\ccalC} +  \mu\bbU_t))]$, but a \emph{biased} estimator of the true reward gradient (when $\mu = 0$) and has high variance (the variance explodes as $\mu$ increases to 0)~\cite{berahas2022theoretical}.
To eliminate this problem, we can use instead a two-point gradient estimator given by
\begin{equation}
    \label{eq:2point-estimate}
    \hat{\nabla} r(\hbG_t^\ccalC) = \frac{\phi(d)}{\mu} \, \left[r\left( \hat{\bbG}_0^{\ccalC}(\hbG_{t}^{\ccalC} +  \mu\bbU_t) \right)-r\left( \hat{\bbG}_0^{\ccalC}(\hbG_{t}^{\ccalC})\right)\right] \cdot \bbU_t,
\end{equation}
which is used in practice in general.
For cases where $r(.)$ is differentiable, the estimator in~\eqref{eq:2point-estimate} is unbiased w.r.t. true gradient (under the assumption that $\mathbb{E}_{\bbU_t \sim \mathcal{D}} [\bbU_t]=0$ and when $\mu \rightarrow 0$.

\paragraph{Best-of-\(N\) direction (greedy ZO).}
Instead of sampling a single direction, this method samples \( N \) candidate directions \( \{\bbU_t^{(1)}, \dots, \bbU_t^{(N)}\} \sim \mathcal{N}(\bbzero, \mathbf{I}) \), and chooses the one that maximizes the reward after denoising:
\begin{equation}
    \label{eq:best_of_n}
    \bbU_t^{(i)} = \argmax_{\{\bbU_t^{(1)}, \dots, \bbU_t^{(N)}\} } \,  r\left(  \hat{\bbG}_0^{\ccalC}(\hbG_{t}^{\ccalC} +  \mu\bbU_t) \right) \cdot \bbU_t.
\end{equation}
The final control is then set as \( \bbU_t = k \cdot \bbU_t^{(i)}  \), where \( k \) is a step size or scaling factor. 
This corresponds to using \( w(\bbU_t) = \mathbbm{1}(\bbU_t = \bbU_t^{(i)} ) \) in~\eqref{eq:zo_general_form}, where $\mathbbm{1}$ represents the indicator function. 
While this method introduces bias, it often leads to effective and low-variance updates, especially when \( r(\cdot) \) is highly non-smooth or sparse.

\paragraph{Multi-point gradient estimator (averaged random search).}
This variant also samples \( N \) directions \( \{\bbU_t^{(1)}, \dots, \bbU_t^{(N)}\} \sim \mathcal{N}(\bbzero , \mathbf{I}) \), but instead of selecting the best, it forms a weighted average of all directions using their corresponding reward evaluations
\begin{equation}
    \label{eq:multipoint}
    \hat{\nabla} r(\hbG_t^\ccalC ) = \frac{1}{N\mu} \sum_{i=1}^N  \, \left[r\left( \hat{\bbG}_0^{\ccalC}(\hbG_{t}^{\ccalC} +  \mu\bbU_t^{(i)}) \right) -r\left( \hat{\bbG}_0^{\ccalC}(\hbG_{t}^{\ccalC})\right)\right]\cdot \bbU_t^{(i)}.
\end{equation}
%
This approach reduces variance compared to both one-point and two-point estimators while maintaining approximate unbiasedness. 
It is especially useful when the reward landscape is moderately smooth, enabling the use of reward information from all sampled directions.

We defer for a quantitative analysis of variance and performance of the three estimators to Appendix~\ref{app:zo-optimization}.  
Overall, these estimators offer flexible trade-offs between estimator quality and query complexity. 
In our setting, we find that the best-of-\(N\) direction yields superior performance in discrete and non-differentiable environments, typical of graph-based objectives.



\paragraph{Final algorithm.} We put everything together and show our proposed algorithm in Alg.~\ref{alg:posterior_opt}.

\begin{algorithm}[t]
    \small
	\caption{GGDiff for controllable generation on graphs}\label{alg:posterior_opt}
	\begin{algorithmic}[1]
		\Require $T, \bbepsilon_{\bbtheta}(\bbG_t, t), N, k, \mu, \{\alpha_t\}_{t=0}^T, \{\sigma_t\}_{t=0}^T, r(\cdot)$ 
        \vspace{0.04in}
        \State Sample $\bbG_T^{\ccalC}$ from $p(\bbG_T)$.
        \vspace{0.04in}
		\For{$t = T-1\; \text{to}\;  1$}
            \State $\hbG_t^\ccalC = \frac{1}{\sqrt{\alpha_{t+1}}} \left( \bbG_{t+1}^\ccalC - \frac{1-\alpha_{t+1}}{\sqrt{1 - \bar{\alpha}}_{t+1}} \bbepsilon_\bbtheta (\bbG_{t+1}^\ccalC, t+1)\right)$ (DDPM update).
            \If{$r$ is differentiable}
                \State Compute $\hbG_0^{\ccalC} ( \hbG_t^\ccalC ) = \frac{1}{\alpha_t}\left(\hbG_t^\ccalC + \sigma_t^2 \bbepsilon_\bbtheta (\hbG_t^\ccalC, t)\right)$.
                \State Compute $\bbU_t = \nabla_{\hbG_t^{\ccalC}} r(\hat{\bbG}_0^{\ccalC}(\hbG_t^{\ccalC}))$ using~\eqref{eq:greedy_approx_control_final_direct}.
            \Else
                \State Sample $N$ candidates $\{\bbU_t^{(1)}, \dots, \bbU_t^{(N)}\} \sim \mathcal{N}(\bbzero , \mathbf{I})$.
                \State Compute $\tilde{\bbG}_t^{\ccalC, (i)} = \hbG_t^\ccalC + k \bbU_t^{(i)}$ for $i=1,\cdots, N$.
                \State Compute $\hbG_0^{\ccalC, (i)} = \frac{1}{\alpha_t}\left(\tbG_t^{\ccalC, (i)} + \sigma_t^2 \bbepsilon_\bbtheta (\tbG_t^{\ccalC, (i)}, t)\right)$ for $i=1,\cdots, N$.
                \State Approximate $\hat{\nabla} r(\hbG_t^\ccalC)$ using~\eqref{eq:2point-estimate}, \eqref{eq:best_of_n}, \eqref{eq:multipoint}
                \If {Approximation of $\hat{\nabla} r(\hbG_t^\ccalC)$ is~\eqref{eq:best_of_n}}
                \State Set $\bbU_t = \argmax_{\bbU_t^{(i)}} r(\hbG_0^{\ccalC, (i)})$.
                \ElsIf {Approximation of $\hat{\nabla} r(\hbG_t^\ccalC)$ is~\eqref{eq:2point-estimate} or~\eqref{eq:multipoint}}
                \State Set $\bbU_t = \hat{\nabla} r(\hbG_t^\ccalC)$.
                \EndIf
            \EndIf
            \State $\bbG_t^\ccalC = \hbG_t^\ccalC + k \bbU_t$.
		\EndFor \\
	\Return $\bbG_{0}^\ccalC$
	\end{algorithmic}
\end{algorithm}
%

\section{Experiments}
\label{sec:experiments}

In our experiments, we evaluate the efficacy of the proposed Graph Guided Diffusion (GGDiff) framework for conditional graph generation across various tasks and reward functions.
We compare the performance of the proposed guidance strategies within the GGDiff framework: ``GGDiff-G'', which is the gradient-based guidance framework for differentiable rewards; ``GGDiff-C'', implementing the greedy Best-of-$N$ approach approach for non-differentiable rewards; and ``GGDiff-Z'', employing the multi-point gradient estimator. As baselines, we also report metrics for PRODIGY~\citep{sharma2024diffuse}, a state-of-the-art method for conditional graph generation, and results from unconstrained graph generation to highlight the impact of guidance.

Our experiments cover three main areas: constrained graph generation (Section~\ref{sub:constrained}), where we assess the ability of GGDiff to generate graphs satisfying specific structural properties and constraints; fair graph generation (Section~\ref{sub:fair_graph}), evaluating the framework's performance in generating graphs that adhere to fairness criteria; and link prediction (Section~\ref{sub:incomplete_graph}, where we aim to generate graphs consistent with partially observed adjacency matrices.
The performance of PRODIGY will only be assessed in the first set of experiments (Section~\ref{sub:constrained}), as it is unable to handle more complex guidance scenarios like the ones presented in Sections~\ref{sub:fair_graph} and~\ref{sub:incomplete_graph}.
Further experimental details, along with more experiments, can be found in Appendix~\ref{app:exp_details}.

\subsection{Constrained Graph Generation}\label{sub:constrained}

We first evaluate GGDiff's performance on constrained graph generation tasks, replicating the experimental setup from the PRODIGY paper~\citep{sharma2024diffuse} to enable direct comparison. 
For this set of experiments, we impose constraints on the maximum degree, edge count, and maximum number of triangles of the generated graphs, on the ego small, community small, and enzymes datasets, described in Appendix~\ref{app:exp_details}. 
To evaluate performance, we use two key metrics: $\Delta$ MMD, which is the metric utilized to assess PRODIGY's performance and measures the difference between the MMD values of the unconstrained dataset and the constrained generated graphs (higher values indicate that the generated graphs are closer to the original data distribution), and Val$_{\mathcal{C}}$, representing the fraction of generated graphs that successfully fulfill the imposed constraint (higher values indicate better constraint adherence).

The results for this set of experiments are presented in Table~\ref{tab:constrained}. 
They demonstrate that our GGDiff methods generally achieve superior performance compared to baselines. Specifically, GGDiff variants tend to exhibit higher $\Delta$ MMD values while also showing higher Val$_{\mathcal{C}}$ scores, demonstrating their capability to satisfy structural constraints without deviating significantly from the prior distribution of the datasets.

\begin{table}[ht!]
\caption{Metrics comparison across datasets and constraints.}
\label{tab:constrained}
\centering
\scalebox{0.9}{\begin{tabular}{cccccccccc}
\toprule
\multirow{2}{*}{\textbf{Constraint}} & \multirow{2}{*}{\textbf{Method}} & \multicolumn{2}{c}{\textbf{Ego Small}} & \multicolumn{2}{c}{\textbf{Community Small}} & \multicolumn{2}{c}{\textbf{Enzymes}} \\
\cmidrule(r){3-4} \cmidrule(r){5-6} \cmidrule(r){7-8}
& & $\Delta$ MMD $\uparrow$ & Val$_{\mathcal{C}}$ $\uparrow$ & $\Delta$ MMD $\uparrow$ & Val$_{\mathcal{C}}$ $\uparrow$ & $\Delta$ MMD $\uparrow$ & Val$_{\mathcal{C}}$ $\uparrow$ \\
\midrule
\multirow{5}{*}{\shortstack{Max\\Degree}} & GGDiff-G & 0.11 & 0.87 & -0.54 & 0.95 & -0.37 & 0.98 \\
& GGDiff-C & \textbf{0.15} & \textbf{0.90} & -0.73 & \textbf{1.00} & -0.39 & \textbf{1.00} \\
& GGDiff-Z & 0.08 & 0.86 & -0.26 & 0.78 & -0.36 & 0.89 \\
& PRODIGY & 0.09 & 0.64 & \textbf{-0.16} & 0.98 & \textbf{0.07} & 0.95 \\
& Uncons. & 0.00 & 0.33 & 0.00 & 0.42 & 0.00 & 0.08 \\
\midrule
\multirow{5}{*}{\shortstack{Edge\\Count}} & GGDiff-G & -0.07 & \textbf{0.91} & -0.33 & 0.84 & -0.47 & \textbf{1.00} \\
& GGDiff-C & 0.27 & 0.63 & \textbf{-0.17} & 0.91 & -0.29 & 0.94 \\
& GGDiff-Z & \textbf{0.28} & 0.67 & -0.38 & 0.73 & -0.12 & 0.69 \\
& PRODIGY & 0.27 & 0.70 & -0.39 & \textbf{1.00} & \textbf{-0.10} & \textbf{1.00} \\
& Uncons. & 0.00 & 0.16 & 0.00 & 0.20 & 0.00 & 0.09 \\
\midrule
\multirow{5}{*}{\shortstack{Triangle\\Count}} & GGDiff-G & \textbf{0.03} & \textbf{0.96} & -0.31 & 0.95 & -0.03 & 0.98 \\
& GGDiff-C & 0.01 & 0.89 & -1.00 & \textbf{1.00} & -0.01 & \textbf{1.00} \\
& GGDiff-Z & -0.07 & 0.88 & -0.14 & 0.85 & -0.04 & \textbf{1.00} \\
& PRODIGY & -0.01 & 0.52 & \textbf{-0.13} & 0.72 & \textbf{0.17} & 0.94 \\
& Uncons. & 0.00 & 0.62 & 0.00 & 0.19 & 0.00 & 0.50 \\\bottomrule
\end{tabular}}
\end{table}

\begin{figure}[t]
    \centering
    \begin{subfigure}{0.24\textwidth}
        \includegraphics[width=\linewidth]{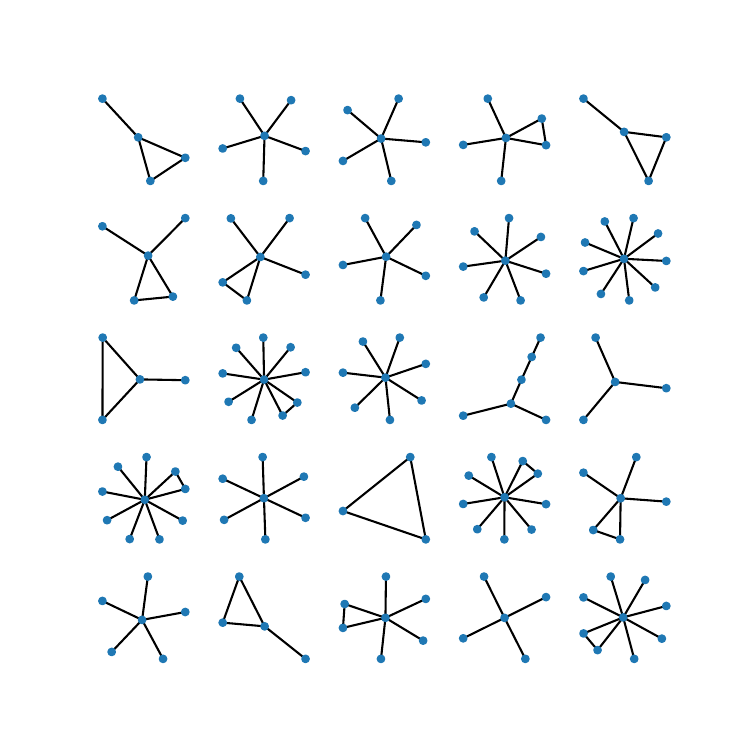}
        \caption{GGDiff-C.}
    \end{subfigure}
    \begin{subfigure}{0.24\textwidth}
        \includegraphics[width=\linewidth]{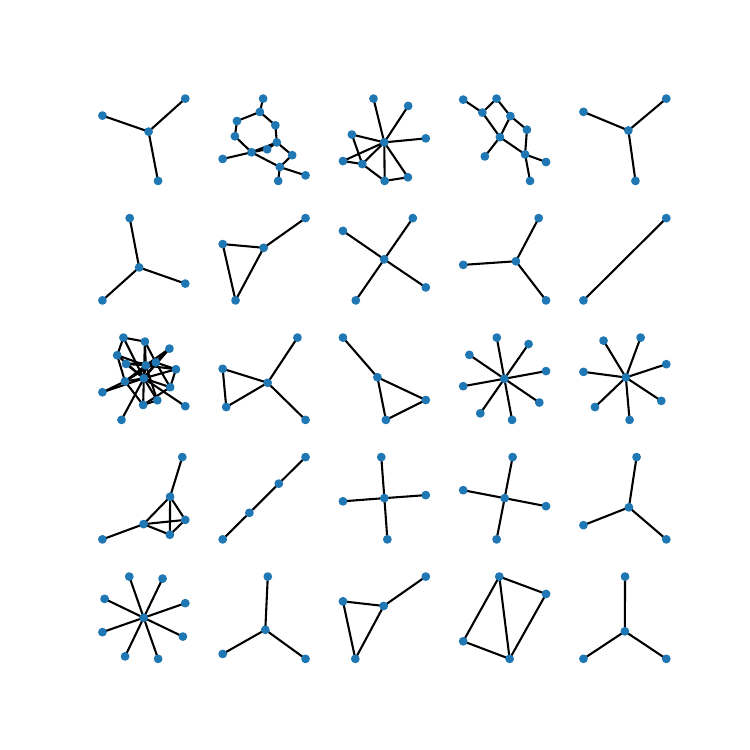}
        \caption{GGDiff-G.}
    \end{subfigure}
    \begin{subfigure}{0.24\textwidth}
        \includegraphics[width=\linewidth]{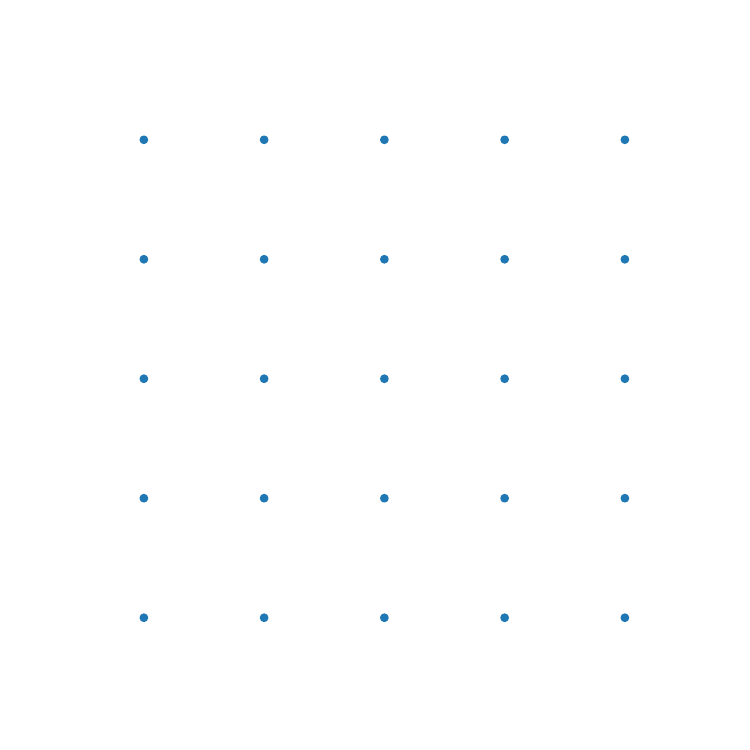}
        \caption{PRODIGY.}
    \end{subfigure}
    \begin{subfigure}{0.24\textwidth}
        \includegraphics[width=\linewidth]{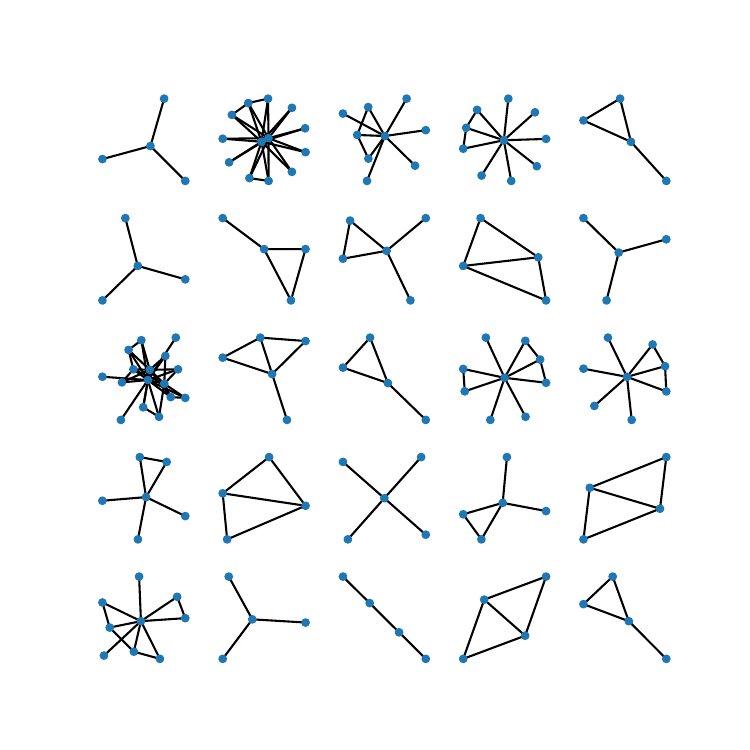}
        \caption{Unconstrained.}
    \end{subfigure}
    \caption{Samples for the force stars constraint in the Ego small dataset.}
    \label{fig:samples_force_stars}
    \vspace{-4mm}
\end{figure}

Moving beyond the constraints explored in~\citet{sharma2024diffuse}, we investigate GGDiff's ability to generate star graphs within the Ego small dataset.
As directly enforcing a star graph structure is outside the standard constraints that can be achieved via projection, for PRODIGY we proxy this by setting the number of triangles to 0, a necessary condition for star graphs.
The results are detailed in Table~\ref{tab:force_stars}, where we report the percentage of graphs with 1 node, the percentage of generated graphs that are stars, the percentage of stars with more than one node, the percentage of valid egonets, and the difference in the number of edges with respect to a star graph, i.e., the ratio between the number of generated graphs that fulfill the condition (having one node, valid egonet, etc.) and the total number of generated graphs.
Our findings indicate that PRODIGY generates graphs consisting of only a single node, as it can be appreciated in Figure~\ref{fig:samples_force_stars}. In contrast, our GGDiff methods successfully double the percentage of generated star graphs compared to the unconstrained case, while effectively preserving the overall data distribution, as approximately 90\% of the graphs generated by GGDiff are valid egonets. Notice that all graphs generated by PRODIGY are valid egonets because a graph with a single node is considered a valid egonet. 
Additionally, our methods substantially reduce the number of excess edges beyond what is required for a star graph over the unconstrained case.

\begin{table}[t]
\caption{Metrics for the force stars constraint in the Ego small dataset.}
\label{tab:force_stars}
\centering
\scalebox{0.9}{\begin{tabular}{cccccc}
\toprule
\textbf{Method} & \textbf{\% 1 Node} & \textbf{\% Stars} & \textbf{\% Stars \& > 1 Node} & \textbf{\% Valid Egonet} & \textbf{Edges over Star} \\
\midrule
GGDiff-G & 0.78 & 53.12 & 52.34 & 96.09 & 1.08 $\pm$ 2.61 \\
GGDiff-L & 2.34 & 51.56 & 49.22 & 88.28 & 0.44 $\pm$ 0.58 \\
PRODIGY & 100.00 & 100.00 & 0.00 & 100.00 & 0.00 $\pm$ 0.00 \\
Uncons. & 0.78 & 24.22 & 23.44 & 99.22 & 1.86 $\pm$ 2.64 \\
\bottomrule
\end{tabular}}
\vspace{-5mm}
\end{table}

\subsection{Fair Graph Generation}\label{sub:fair_graph}

\begin{wraptable}{r}{0.68\textwidth}
\centering
\caption{Metrics for the fair graph generation experiment.}
\label{tab:fair_graph_metrics}
\scalebox{0.95}{\begin{tabular}{ccccc}
\toprule
\textbf{Method} & \textbf{$\Delta$ DP} & \textbf{$\Delta \text{DP}_{\text{node}}$ } & \textbf{\% Valid SBM} \\
\midrule
GGDiff-G & 0.0026 \footnotesize{$\pm$ 0.0029} & 0.0249 \footnotesize{$\pm$ 0.0125} & 100.0000 \\
GGDiff-C & 0.0035 \footnotesize{$\pm$ 0.0053} & 0.0192 \footnotesize{$\pm$ 0.0121} & 99.2188 \\
GGDiff-Z & 0.0015 \footnotesize{$\pm$ 0.0020} & 0.0061 \footnotesize{$\pm$ 0.0037} & 95.3125 \\
Uncons. & 0.0071 \footnotesize{$\pm$ 0.0145} & 0.0295 \footnotesize{$\pm$ 0.0218} & 99.2188 \\
\bottomrule
\end{tabular}}
\end{wraptable}

In this section, we evaluate GGDiff's performance on generating fair graphs using metrics defined in \citet{navarro24fairglasso}. For these experiments, we randomly assign sensitive attributes to the nodes of the graphs generated from the community small dataset (for a similar experiment where the communities of the nodes are assigned by a community detection algorithm, refer to Appendix~\ref{app:exp_details}).
We report two key fairness metrics from \citet{navarro24fairglasso}: $\Delta$ DP and $\Delta$ DP$_{\text{node}}$, where lower values indicate greater dyadic parity and thus fairer graphs.
To assess whether the generated graphs maintain the underlying SBM structure of the dataset, we report the percentage of valid SBMs. An SBM is considered valid if its estimated intra-community edge probability is at least 8 times its inter-community edge probability; this factor was chosen such that 95\% of the test graphs in the dataset fulfill this criterion.

The results in Table~\ref{tab:fair_graph_metrics} demonstrate that our GGDiff methods effectively reduce the fairness metrics ($\Delta$ DP and $\Delta$ DP$_{\text{node}}$) and increase the number of edges between nodes with different sensitive attributes, indicating improved fairness.
Crucially, these improvements are achieved while largely maintaining the generated graphs within the family of the prior distribution (SBMs), as reflected in the percentage of valid SBMs.

\subsection{Incomplete Graph Generation}\label{sub:incomplete_graph}

In this task, we evaluate GGDiff's ability to generate graphs consistent with partially observed adjacency matrices. Specifically, we assume that 50\% of the entries of the adjacency matrix are observed and should be maintained in the generated graph. We conduct these experiments on two molecular datasets, QM9 and ZINC250k. We evaluate performance using three metrics: Accuracy, which measures the percentage of observed entries that are respected in the generated graphs; and \% Unique, the percentage of generated molecules that are novel compared to the training set.

\begin{wraptable}{r}{0.63\textwidth}
\centering
\caption{Results for the incomplete graph generation experiment.}
\label{tab:incomplete_graph_gen}
\scalebox{0.95}{
\begin{tabular}{ccccccc}
\toprule
\multirow{2}{*}{\textbf{Method}} & \multicolumn{2}{c}{\textbf{QM9}} & \multicolumn{2}{c}{\textbf{ZINC250k}} \\ \cmidrule(r){2-3} \cmidrule(r){4-5}
& Acc. (\%) & \% Unique & Acc. (\%) & \% Unique \\
\midrule
GGDiff-G & 91.39 & 73.57 & 98.85 & 100.00 \\
GGDiff-C & 67.20 & 94.32 & 95.27 & 100.00 \\
GGDiff-Z & 88.72 & 90.88 & 98.44 & 100.00 \\
Uncons. & 61.51 & 97.86 & 93.43 & 100.00 \\
\bottomrule
\end{tabular}
}
\end{wraptable}

The results are presented in Table~\ref{tab:incomplete_graph_gen}. Our GGDiff methods, particularly GGDiff-G and GGDiff-Z, demonstrate high accuracy in respecting the observed entries, both of them achieving almost 90\% accuracy on QM9 and over 98\% accuracy on ZINC250k. The unconstrained case also shows relatively high accuracy, which is largely attributable to the high prevalence of zero entries (absence of edges) in sparse graphs. For a more challenging evaluation where we observe edges instead of entries, please refer to Appendix~\ref{app:exp_details}. 
Across all methods, the percentage of valid generated molecules is consistently 100\%, likely aided by the partial observation of the adjacency matrix.
We observe a trade-off between accuracy and novelty in the QM9 dataset: as the accuracy in fixing observed entries increases, the percentage of novel molecules tends to decrease, suggesting that achieving very high fidelity to observed structure can lead to generating molecules highly similar to those in the test set. This tradeoff isn't observed in the ZINC250k dataset, likely due to the fact that the graphs are larger and therefore the model has more freedom to adapt to the observed entries.

\section{Conclusions}
\label{sec:concl}

In this paper, we introduced \textbf{Graph Guided Diffusion (GGDiff)}, a flexible, gradient-free framework for conditional graph generation, grounded in stochastic optimal control. 
By casting guidance as a control problem, GGDiff enables plug-and-play conditioning of pre-trained diffusion models under both differentiable and black-box constraints.
GGDiff unifies a range of existing guidance approaches, including gradient-based guidance and non-differentiable cases, under a single SOC-based formulation. 
Our method supports both hard and soft constraints without requiring gradient access or projection operators, making it broadly applicable across domains.
Extensive experiments on structural, fairness, and topology-based constraints demonstrate GGDiff's effectiveness and generality, outperforming prior work in handling complex, non-differentiable objectives.

Despite its flexibility, GGDiff has limitations. 
ZO optimization adds computational overhead and may suffer from high variance in challenging settings. Future work includes integrating GGDiff with discrete diffusion models to better handle discrete constraints, and exploring more efficient or adaptive ZO gradient estimators to improve scalability and performance.

\section*{Acknowledgments}

This research was sponsored by the Spanish AEI (10.13039/501100011033) under Grants PID2022-136887NBI00 and FPU20/05554; by the Community of Madrid within the ELLIS Unit Madrid framework and the IDEA-CM (TEC-2024/COM-89) and CAM-URJC F1180 (CP2301) grants; by the Army Research Office under Grant Number W911NF-17-S-0002; by the National Science Foundation under award CCF-2340481 and by a Ken Kennedy Institute 2024–25 Ken Kennedy-HPE Cray Graduate Fellowship. The views and conclusions contained in this document are those of the authors and should not be interpreted as representing the official policies, either expressed or implied, of the Army Research Office, the U.S. Army, or the U.S. Government. The U.S. Government is authorized to reproduce and distribute reprints for Government purposes, notwithstanding any copyright notation herein.

\bibliographystyle{apalike}
\bibliography{citations}


\appendix

\section{HJB equation}
\label{app:oc_theory}

In this section, we give more details on our SOC formulation.
The optimal control is given by
\[
\bbU^*(\bbG_t^\ccalC, t) = -\frac{g(t)}{\lambda}\nabla_{\bbG_t^\ccalC} V_t^*\left(\bbG_t^\ccalC \right)
\]
where \( V_t^*(\bbG_t^\ccalC ) \) is the optimal value function~\cite{pavon1989stochastic}.  
For our problem, the optimal value function at time $t$ is given by
\begin{equation}
V_t^*(\bbG_t^\ccalC ) = \mathbb{E}_{p_t^*} \left[ \int_t^0 \lambda\frac{\|
\bbU^*(\bbG_s^\ccalC , s)\|_2^2}{2} \mathrm{d}s - r(\bbG_0^\ccalC ) \,\Big|\, \bbG_t^\ccalC \right]
\label{eq:value_function}
\end{equation}
where \( p_t^* \) denotes the optimal \emph{controlled} distribution at time $t$ given by $p_t^*(\bbG) \propto  \exp\left( \frac{-V_t^*(\bbG)}{\lambda} \right) p_t^{\mathrm{pre}}(\bbG)$ and $p_t^{\mathrm{pre}}$ is the prior (uncontrolled) distribution\footnote{We assume here that the terminal time is $0$ and the time runs backwards (so $t < 0$).}.
The value function \( V_t^* \) solves the stochastic Hamilton-Jacobi-Bellman (HJB) equation~\cite{evans2022partial}, given by
\begin{align}
    &\partial_t V_t^*(\mathbf{G}_t^{\ccalC}) = \\\nonumber
    & +
    \left(\nabla_{\mathbf{G}_t^{\ccalC}} V_t^*(\mathbf{G}_t^{\ccalC})\right)^T \bbmu(\bbG_t^{\ccalC}, t)
    -
    \frac{g(t)^2}{2\lambda} \left\|\nabla_{\mathbf{G}_t^{\ccalC}} V_t^*(\mathbf{G}_t^{\ccalC})\right\|_2^2 
    +
    \frac{1}{2} g(t)^2 \Delta_{\mathbf{G}_t^{\ccalC}} V_t^*(\mathbf{G}_t^{\ccalC}),
\end{align}
with boundary condition $V_0(\bbG_0^{\ccalC}) = r(\bbG_0^{\ccalC})$, and
where $\bbmu(\bbG_t^{\ccalC}, t) = \frac{1}{2}\mathbf{G}_t^{\ccalC} - g(t)^2\nabla_{\mathbf{G}_t^{\ccalC}}\log p(\mathbf{G}_t^{\ccalC})$.

This equation is a \emph{non-linear} partial differential equation (PDE), and the solution to the non-linear HJB equation is nontrivial.
However, by applying an exponential transformation $\phi_t(\bbG_t^{\ccalC}) = e^{-V_t(\bbG_t^{\ccalC})}$,  we can obtain the \emph{linear} HBJ equation, given by
\begin{equation}
    \label{eq:linear_HJB}
    -\partial_t \phi(\mathbf{G}_t^{\ccalC}, t) = 
    \left(\nabla_{\mathbf{G}_t^{\ccalC}} \phi(\mathbf{G}_t^{\ccalC}, t)\right)^T \bbmu(\bbG_t^{\ccalC}, t)
    +
    \frac{1}{2} g(t)^2 \Delta_{\mathbf{G}_t^{\ccalC}} \phi(\mathbf{G}_t^{\ccalC}, t)
\end{equation}
In particular, the Feynman-Kac formula is obtained as the solution of the linearized HJB equation in~\eqref{eq:linear_HJB} (see~\cite{oksendal2003stochastic} for the proof), given by
\begin{equation}
\label{eq:feynman-kac}
\exp\left( \frac{V_t^*(\bbG)}{\lambda} \right) = \mathbb{E}_{p^{\mathrm{pre}}} \left[ \exp\left( \frac{-r(\bbG_0^\ccalC )}{\lambda} \right) \,\Big|\, \bbG_t^\ccalC = \bbG \right].
\end{equation}
This leads to an expression for the optimal control in terms of the reward function as given by~\eqref{eq:optimal_control}.

\paragraph{Stochastic optimal control for zero-shot controlled generation.} 
Recent methods have proposed the use of SOC for controlled generation~\cite{uehara2025inference, li2024derivative}.
In the context of music generation~\cite{huang2024symbolic}, the authors propose a method to generate samples when likelihoods are non-differentiable.
In~\cite{rout2024solving}, a linear quadratic control was proposed for style transfer in image generation.
More recently, a non-linear control formulation was introduced in~\cite{pandey2025variational} for image inverse problems.
However, as far as we are concerned, the application of SOC for graph generation has not been explored yet.

\section{Background on zeroth-order optimization}
\label{app:zo-optimization}

In Section~\ref{subsubsec:zero_order}, we leverage zeroth-order optimization for defining a surrogate gradient of the reward function.
We propose three estimators in particular, where each one has its own properties.
In this section, we expand on them.

\paragraph{Two-point gradient estimator.} 
The two-point gradient estimator in~\eqref{eq:2point-estimate} is the first one that we introduced.
This estimator has a mean-squared error given by
\begin{equation}
    \label{eq:2point-error}
    \mathbb{E}[\|\hat{\nabla} r(\bbG_0) - \nabla r((\bbG_0))\|_2^2] = O(d)\|\nabla r(\bbG_0)\|_2^2 + O\left(\frac{\mu^2 d^3 + \mu^2 d}{\phi(d)}\right)
\end{equation}
The proof can be found in~\cite{liu2018zeroth}.
The error in~\eqref{eq:2point-error} sheds light on the behavior of this estimator.
First, the second term depends on the parameter $\bbmu$: when this parameter gets smaller, the gradient estimate gets better.
However, if $\bbmu$ becomes too small, then the effect of the guidance diminishes.
Second, the first term depends on the dimension $d$.
This imposes a variance which cannot be 0 even for small values of $\mu$.

\paragraph{Multi-point gradient estimator.} 
The third estimator is based on the multi-point gradient estimate, which computes an average over random directions.
This estimator has a mean-squared error given by
\begin{equation}
    \label{eq:multipoint-error}
    \mathbb{E}[\|\hat{\nabla} r(\bbG_0) - \nabla r((\bbG_0))\|_2^2] = O\left(\frac{d}{N}\right) \|\nabla f(\mathbf{x})\|_2^2 + O\left(\frac{\mu^2 d^3}{\phi(d) N}\right) + O\left(\frac{\mu^2 d}{\phi(d)}\right)
\end{equation}
Compared to the two-point case, the error in~\eqref{eq:multipoint-error} depends on the number of samples that are used to compute the average. 
In particular, the first two terms go to 0 when $N\rightarrow \infty$; the third term is independent of $N$, and corresponds to the approximation error between the true gradient and the smoothed version.
However, it is controlled by the smoothing parameter $\bbmu$.

A summary of each estimator is shown in Table~\ref{tab:zo_comparison}. 

\begin{table}[h]
\centering
\caption{\small Comparison of ZO estimators for control direction optimization.}
\label{tab:zo_comparison}
\renewcommand{\arraystretch}{1.3}
\begin{tabular}{@{} l|c|c @{}}
\toprule
\textbf{Method} & \textbf{Variance} & \textbf{Reward evaluation} \\
\midrule
2-Point Estimator & High & 2 \\
Best-of-\(N\) Direction & Low & \(N\) \\
Averaged Random Search &  Moderate & \(N + 1\) \\
\bottomrule
\end{tabular}
\end{table}


\section{Diffusion models for graph generation.}

Graph diffusion models have been developed in both \emph{continuous} and \emph{discrete} domains. 

\paragraph{Continuous domain.}
The continuous formulation was introduced in EDP-GNN~\citep{niu2020permutation} to diffuse the graph topology, later extended in GDSS~\citep{jo2022score} to include node features, and further explored in the spectral domain~\citep{luo2023fast, minellogenerating}.
Lastly, in~\cite{zhou2024unifying}, the authors proposed to use a latent diffusion model.
The main difference between the latent and the node-level formulation is that the former defines the diffusion process in a latent space.
This requires an encoder-decoder pair to map from the node to the latent space and vice versa.
Formally, the encoder $\bbZ_0 = \ccalE(\bbA)$ maps the adjacency matrix to a latent variable $\bbZ_0$.
Then, $\bbZ_0$ follows a diffusion process similar to the one in Section~\ref{subsec:background_diffusion}, and then a decoder is used to generate an adjacency matrix $\hbA = \ccalD(\bbZ_0)$.
At a high level, continuous models focus on capturing \emph{global structure}

\paragraph{Discrete domain.}
Discrete diffusion was introduced in DiGress~\citep{vignac2023digress} by adapting the structured diffusion framework~\citep{austin2021structured}, framing generation as edge-wise classification to mitigate combinatorial complexity. 

In a nutshell, each node and edge in the graph $\ccalG$ is assumed to take on values from a fixed set of possible categories, and the model learns a categorical distribution for these variables. In other words, the graph represents a multivariate random variable, where each component follows a categorical distribution.
Then, the forward process perturbs the graph by sampling from a modified, discrete, distribution, until it converges to a stationary state (such as a mask or uniform; see~\cite{austin2021structured}).
The “reverse process”, on the other hand, involves recovering the original graph by denoising this uniformly random structure step by step.

While discrete methods are well-suited for sparse graphs, they rely on mean-field approximations and lack gradients, limiting constrained generation. 
To address inference speed, EDGE~\citep{chen2023efficient} proposed using the empty graph as the stationary distribution, achieving faster sampling than DiGress but retaining the gradient limitation.
Given these trade-offs, we focus on the continuous setting (see Appendix~\ref{sec:app_graph_guidance} for details).

\section{Diffusion models for controllable generation on graphs}
\label{sec:app_graph_guidance}

Diffusion models have shown impressive results in unconstrained graph generation, either when considering discrete diffusion~\cite{vignac2023digress,chen2023efficient} or score-based approaches, i.e., considering a continuous relaxation~\citep{niu2020permutation, jo2022score}.
However, only a few works consider the generation of graphs under constraints when using diffusion models as priors.
DiGress~\cite{vignac2023digress} incorporates guidance based on a learned regressor that encodes the requirements in a feature vector.
However, this requires learning an additional model.
LGD~\cite{zhou2024unifying} considers a classifier-free guidance approach~\cite{ho2021classifier} in a latent space.
To incorporate more complex constraints, in~\cite{madeira2024generative} the authors propose to use a projection operator combined with an edge-absorbing model.
While effective, the projection operator is computationally demanding.
PRODIGY~\cite{sharma2024diffuse}, on the other hand, leverages continuous diffusion models and incorporates a projection operator.
While the projection operator can be implemented efficiently using the bisection method, this only holds for simple constraints with a closed-form projection operator. 
This reliance limits its applicability since such operators are often unavailable.
As a result, more complex constraints require computationally intensive solvers, significantly increasing runtime and restricting the scalability of PRODIGY to more general settings.

\section{Experimental Details}
\label{app:exp_details}

This appendix provides detailed information regarding the experimental setup used in this paper, including specifics about the datasets, computational resources utilized, and a comprehensive description of additional experiments conducted. The appendix is structured as follows: Section~\ref{app:datasets} includes a description of the datasets used for evaluating GGDiff's performance. Section~\ref{app:comp_resources}, details the computational resources of the server where the experiments were run. Section~\ref{app:additional_exps} presents additional experimental results, with subsections dedicated to further details on constrained graph generation (Section~\ref{app:extra_constrained}), fair graph generation (Section~\ref{app:extra_fair}) and link prediction (Section~\ref{app:extra_incomplete}).

\subsection{Datasets}\label{app:datasets}

We evaluate our proposed GGDiff framework and baselines on a selection of benchmark graph datasets, encompassing both generic network structures and molecular graphs. The datasets used in our experiments are described below:

\begin{enumerate}
    \item \textbf{Ego-small}: This dataset comprises 200 small ego graphs extracted from the larger Citeseer network. 
    \item \textbf{Community-small}: Consisting of 100 synthetic graphs, this dataset features structures exhibiting distinct community partitions.
    \item \textbf{Enzymes}: We use the protein graphs from the BRENDA enzyme database, totaling 587 graphs. 
    \item \textbf{QM9}: A molecular dataset containing approximately 133,000 small molecules. These molecules are composed of up to 9 heavy atoms, including Carbon (C), Nitrogen (N), Oxygen (O), and Fluorine (F).
    \item \textbf{ZINC250k}: This large molecular dataset includes 250,000 drug-like molecules. The graphs represent molecules with 6 to 38 heavy atoms, incorporating Carbon (C), Nitrogen (N), Oxygen (O), Fluorine (F), Phosphorus (P), Chlorine (Cl), Bromine (Br), and Iodine (I).
\end{enumerate}

\subsection{Computational resources}\label{app:comp_resources}

All experiments were conducted on a server equipped with an AMD EPYC 9634 84-Core Processor and 512GB of total physical memory (RAM). For accelerated computation, the server uses an NVIDIA GeForce RTX 4090 graphics processing units (GPUs), each featuring 24GB of dedicated video memory. The software environment runs on Ubuntu 24.04 LTS, with NVIDIA driver version 560.35.03 and CUDA version 12.6.

\subsection{Additional experiments}\label{app:additional_exps}

\subsubsection{Constrained Graph Generation}\label{app:extra_constrained}

In this section, we provide additional details regarding the constrained graph generation experiments summarized in the main paper (see Table~\ref{tab:constrained}). For comparison purposes with prior work, we specifically focus on evaluating GGDiff's performance on the task of guiding the generated graphs towards fulfilling the constraints previously defined and utilized in \citet{sharma2024diffuse}. These constraints, designed to enforce specific structural properties, are presented in Table~\ref{tab:constraints_appendix}, along with their descriptions and mathematical formulations.

\begin{table}[htbp]
\centering
\caption{Summary of Constraints from \cite{sharma2024diffuse}}
\label{tab:constraints_appendix}
\begin{tabular}{p{0.2\linewidth} | p{0.23\linewidth} | p{0.45\linewidth}}
\toprule
\textbf{Constraint Type} & \textbf{Limiting factor} & \textbf{Mathematical Formulation} \\
\midrule
Edge Count & Number of edges $|\ccalE|$ & $|\ccalE| = \bbone^\top \bbA \bbone \le B$ for a given constant $B \ge 0$ \\
Triangle Count & Number of triangles & $\text{tr}(A^3) \le T$ for a given constant $T \ge 0$ \\
Degree & Maximum Degree & $\max_i [\bbA \mathbf{1}]_i \le D$ for a given constant $D$ \\
\bottomrule
\end{tabular}
\end{table}

The values for constants $B$, $T$, and $D$ used for each dataset are selected based on those reported in \cite{sharma2024diffuse} to ensure a direct comparison of method performance under identical constraint settings, and are given by those values fulfilled by 10\% of the graphs in the test dataset.

The specific loss function used for each constraint is empirically selected from a pool of possibilities based on which yields the best performance; a comprehensive list of options can be found in the code associated with this submission.
For differentiable guidance (Section~\ref{subsubsec:differentiable}), the choice is restricted to differentiable functions, typically involving $\ell_1$ or $\ell_2$ norms. For instance, an $\ell_2$ loss for the edge count constraint could be $(\bbone^\top \hbA_0^\ccalC (\bbA_t^\ccalC) \bbone - B)^2$.
In contrast, the non-differentiable (zero-order) guidance (Section~\ref{subsubsec:zero_order}) significantly expands the available loss functions. Examples include utilizing non-differentiable operations in the differentiable losses, like using the quantized adjacency via the entry-wise indicator function $\mathbbm{1}(\hbA_0^\ccalC (\bbA_t^\ccalC) > 0.5)$ in lieu of the estimate $\hbA_0^\ccalC (\bbA_t^\ccalC)$, or employing one-sided penalties such as $\max \{\bbone^\top \hbA_0^\ccalC (\bbA_t^\ccalC) \bbone - B, 0\}$.

\subsubsection{Fair graph generation}\label{app:extra_fair}

In this appendix section, we provide further details regarding the fair graph generation experiments introduced in the main paper. These experiments evaluate GGDiff's ability to generate graphs that satisfy fairness criteria based on assigned sensitive attributes.
To encourage fair graphs, we employ the same loss functions defined in~\citet{navarro24fairglasso}.
We investigate two distinct methods for assigning sensitive attributes to the nodes of the community small dataset:

\begin{enumerate}
    \item \textbf{Random assignment}: Sensitive attributes are assigned to nodes randomly. The quantitative results for the fairness metrics and SBM validity for this scenario are presented in Table~\ref{tab:fair_graph_metrics} in the main paper.
    \item \textbf{Community partitioning algorithm-based assignment}: Sensitive attributes are assigned to nodes based on the community structure identified by a community partitioning algorithm. This represents a more challenging scenario for generating fair graphs that are also valid Stochastic Block Models (SBMs). Since SBMs are characterized by a high density of intra-community edges and a low density of inter-community edges, aligning the sensitive attribute with community membership creates a direct tension: the fair loss function encourages the formation of edges between nodes with different attributes (i.e., nodes in different communities), while the underlying data distribution and the objective of generating valid SBMs favor the opposite.
\end{enumerate}

\begin{figure}
    \centering
    \begin{subfigure}{\textwidth}
    \centering
    \includegraphics[width=\linewidth]{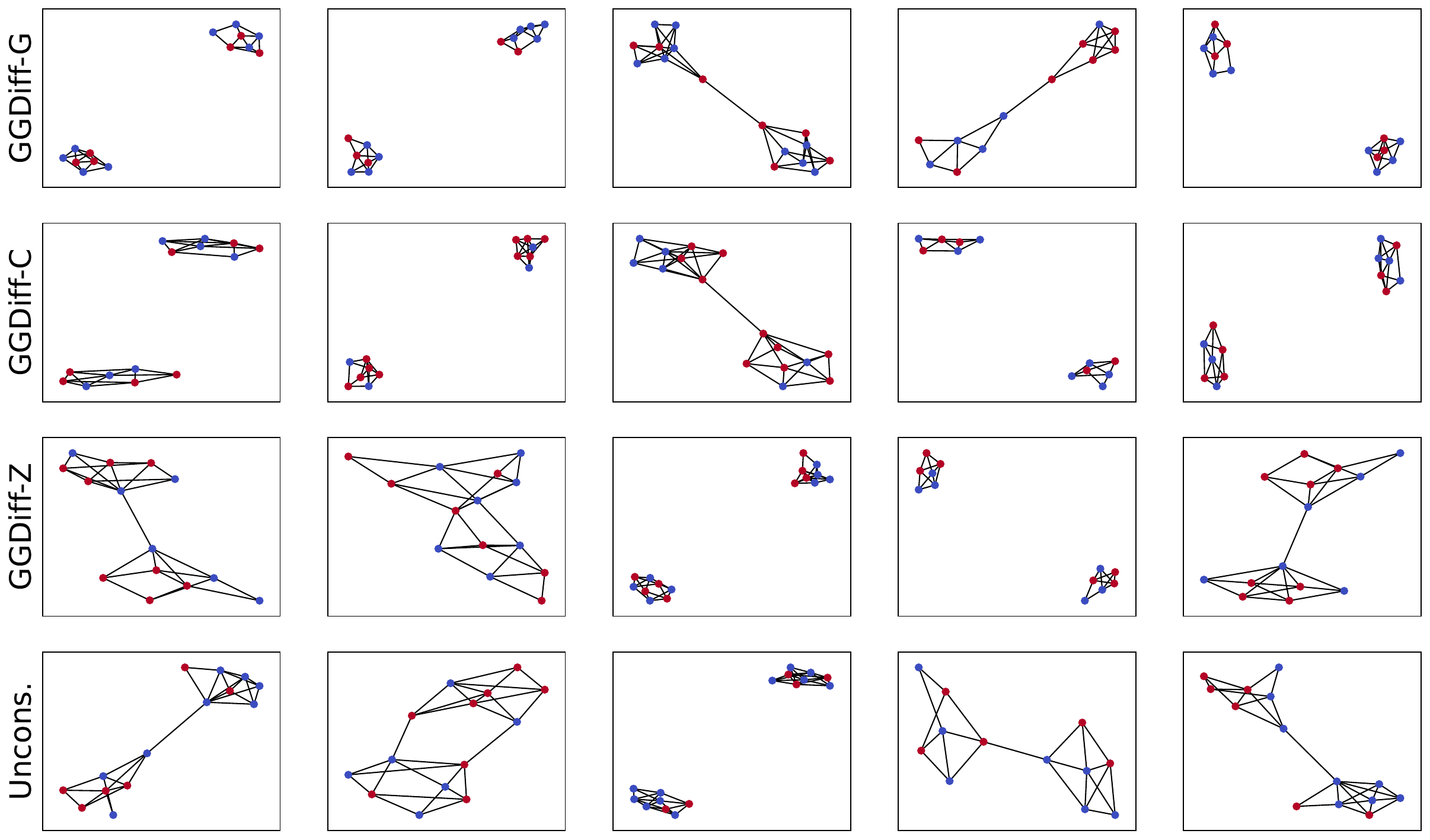}
    \caption{Random community assignment.}
    \label{subfig:fair_random}
    \end{subfigure}
    \noindent\rule{\textwidth}{0.4pt}
    \begin{subfigure}{\textwidth}
    \centering
    \includegraphics[width=\linewidth]{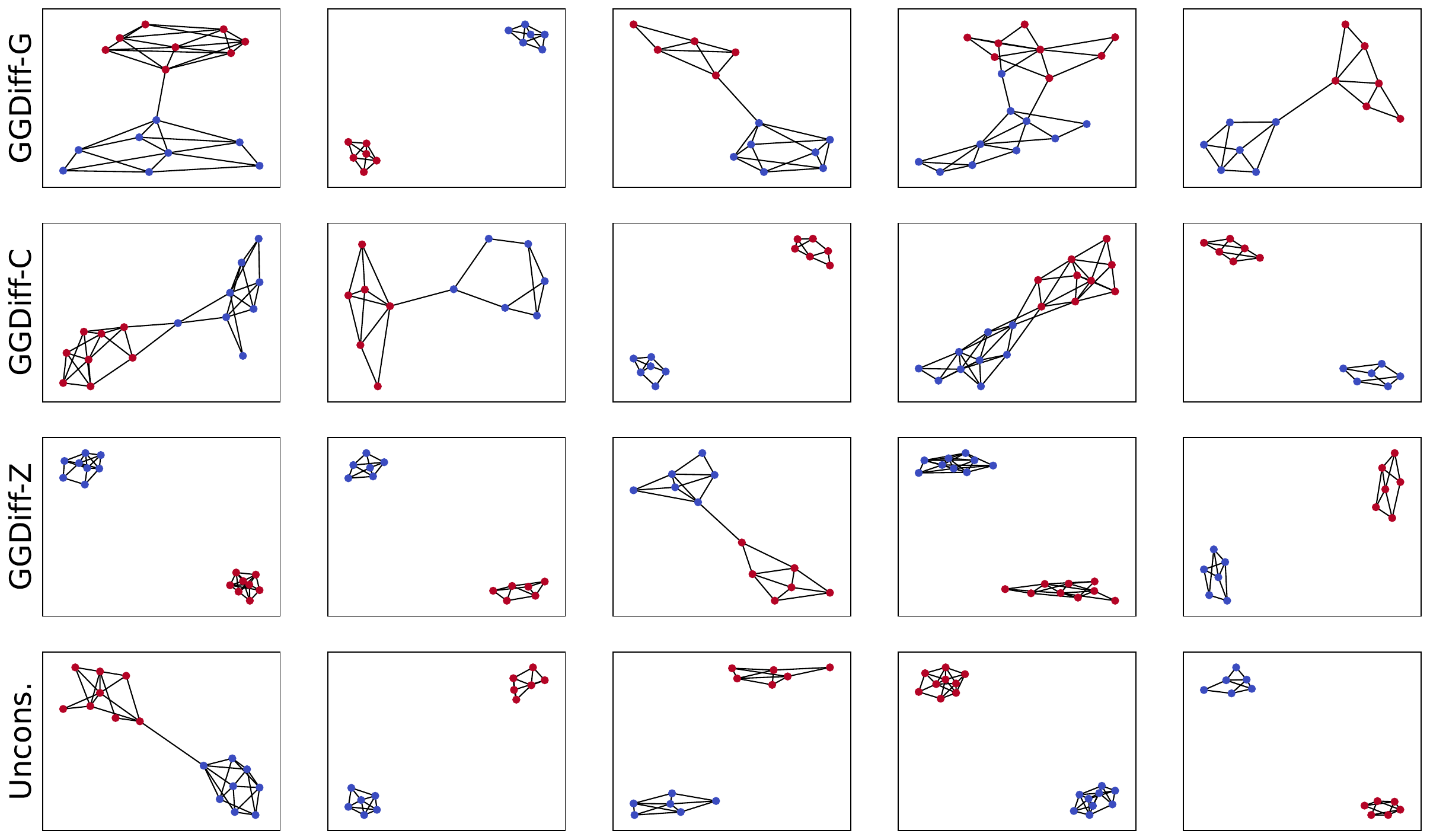}
    \caption{Community partitioning algorithm-based assignment.}
    \label{subfig:fair_comm}
    \end{subfigure}
    \caption{Samples from the fair graph generation experiment.}
    \label{fig:fair_samples}
\end{figure}

Analyzing the results for the community partitioning-based assignment presented in Table~\ref{tab:fair_graph_metrics_community}, our three GGDiff methods are still able to effectively reduce the fairness metrics compared to baselines, while largely maintaining a high percentage of valid SBMs. This quantitative improvement is visually corroborated by the sample graphs shown in Figure~\ref{fig:fair_samples}, where graphs generated by GGDiff show a higher density of edges connecting nodes of different sensitive attributes (indicated by node color) compared to the unconstrained case.

\begin{table}[ht]
\centering
\caption{Metrics for the fair graph generation with community partition.}
\label{tab:fair_graph_metrics_community}
\begin{tabular}{ccccc}
\toprule
\textbf{Method} & \textbf{$\Delta$ DP} & \textbf{$\Delta \text{DP}_{\text{node}}$ } & \textbf{\% Valid SBM} \\
\midrule
Greedy & 0.3389 $\pm$ 0.0763 & 0.1931 $\pm$ 0.0387 & 98.4375 \\
Loss & 0.3119 $\pm$ 0.1289 & 0.1892 $\pm$ 0.0720 & 77.3438 \\
Zero & 0.3451 $\pm$ 0.0612 & 0.1999 $\pm$ 0.0498 & 98.4375 \\
Uncons & 0.4133 $\pm$ 0.0769 & 0.2348 $\pm$ 0.0449 & 99.2188 \\
\bottomrule
\end{tabular}
\end{table}

\subsubsection{Incomplete graph generation}\label{app:extra_incomplete}

This section provides additional details on the link prediction experiments, also referred to as incomplete graph generation. In this task, we evaluate GGDiff's ability to generate graphs where a subset of adjacency matrix entries is observed and must be precisely replicated in the generated output. We investigate two scenarios for the observed entries: (i) observing a random 50\% of all adjacency matrix entries (both existing edges and non-edges), and (ii) observing only a random subset of existing edges (entries equal to 1).
This task is particularly relevant in domains like molecule generation, where there is often an interest in generating molecules that incorporate a specific predefined substructure (e.g., a benzene ring). Enforcing observed edges allows for the generation of molecules that respect such topological constraints.

The results for the first scenario (observing random entries) are presented in Table~\ref{tab:incomplete_graph_gen} in the main paper. As noted, the high accuracy values observed in this case are significantly influenced by the correct generation of prevalent zero entries (non-edges) that were part of the observed subset. We now detail the second, more challenging scenario in this appendix.

For the second scenario, we observe only a random subset of existing edges in the adjacency matrix. The results for this case are presented in Table~\ref{tab:incomplete_graph_gen_edges}. Here, the accuracy metric specifically measures how well the generated graphs reproduce the observed edges. As expected, the accuracy values drop significantly compared to the first scenario because correctly generating existing edges is a more stringent condition than correctly generating non-edges in sparse graphs. However, the effect of GGDiff's guidance becomes strikingly apparent: our methods, particularly GGDiff-G and GGDiff-Z, achieve drastically increased accuracy in reproducing the observed edges on both the QM9 and ZINC250k datasets compared to the unconstrained baseline.

\begin{figure}
    \centering
    \begin{subfigure}{\textwidth}
    \centering
    \includegraphics[width=0.85\linewidth]{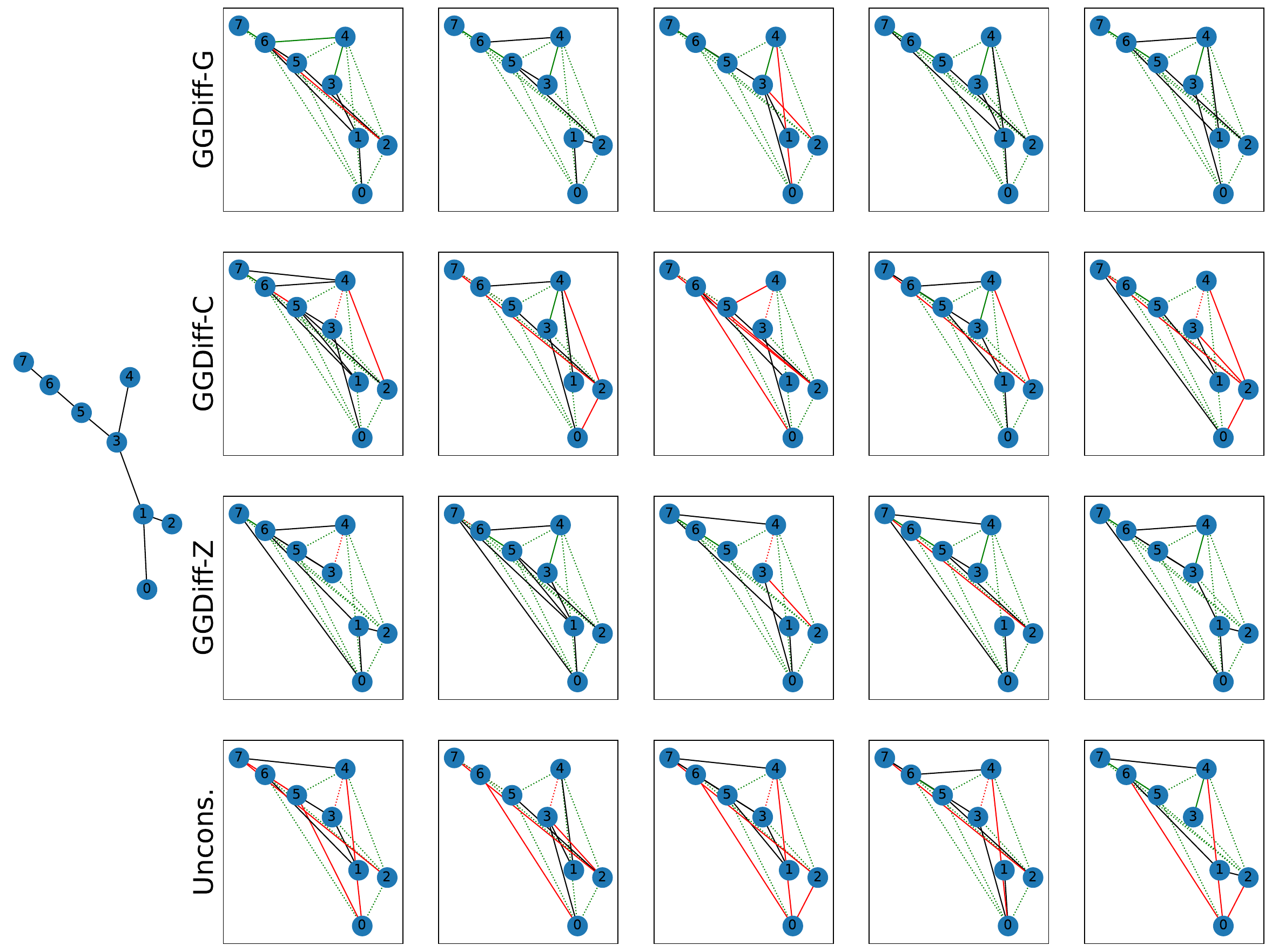}
    \caption{QM9.}
    \end{subfigure}
    \noindent\rule{\textwidth}{0.4pt}
    \begin{subfigure}{\textwidth}
    \centering
    \includegraphics[width=0.85\linewidth]{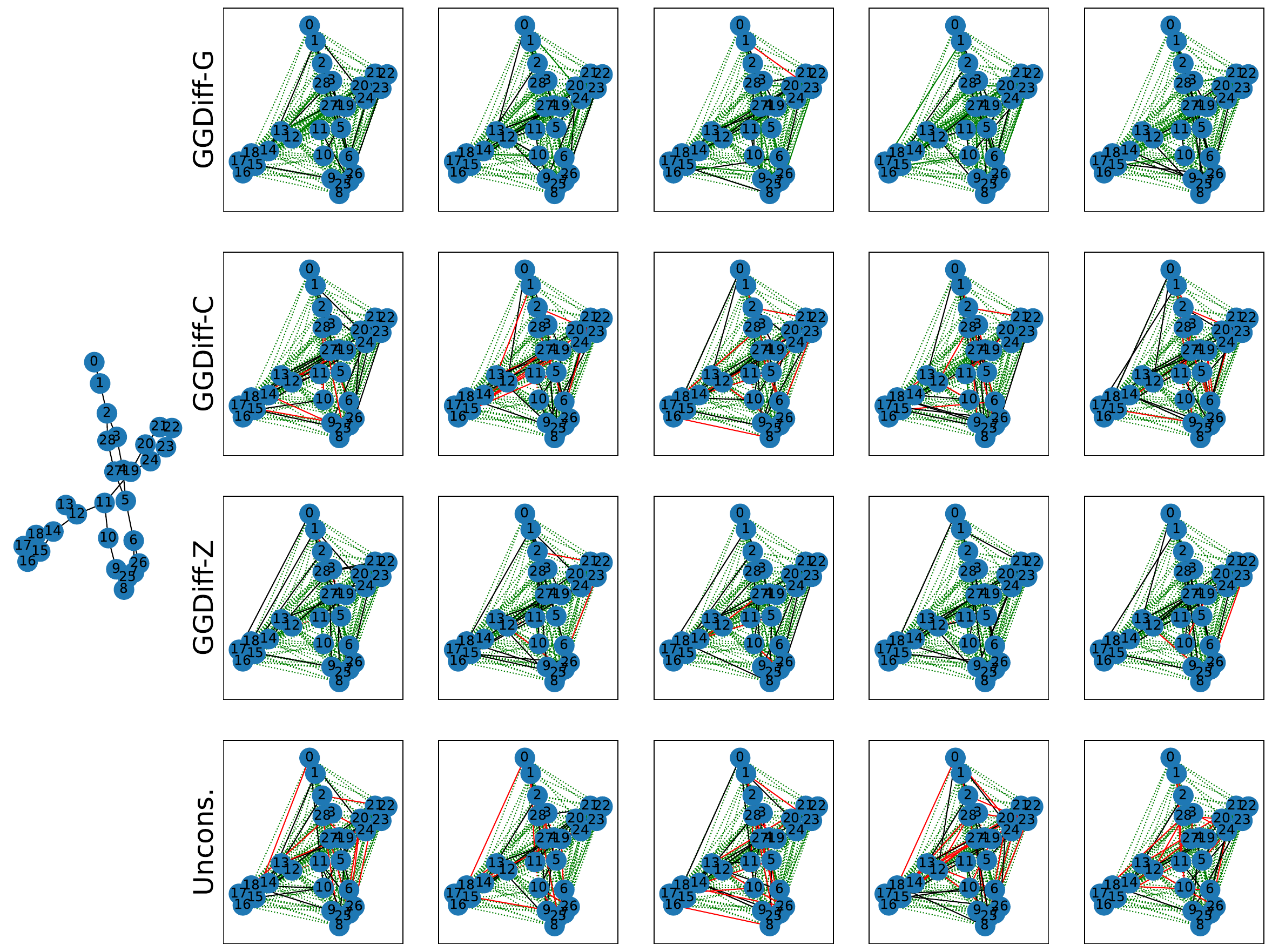}
    \caption{ZINC250k.}
    \end{subfigure}
    \caption{Samples generated for the incomplete graph generation experiment. The graph on the left is the test graph from which we observe the entries in its adjacency matrix. The generated graphs are represented in the rows, one for each of the methods. In the generated graphs, the solid green (red) lines are observed edges that were (not) preserved in the generated graphs, while dotted green (red) lines are observed entries not corresponding to an edge that were (not) preserved in the generated graphs.}
    \label{fig:samples_incomplete}
\end{figure}

Sample graphs illustrating the results of the link prediction experiment are shown in Figure~\ref{fig:samples_incomplete}. In these visualizations, we use color and line style to indicate the status of observed entries in the generated graphs:
\begin{itemize}
    \item \textbf{Solid green lines}: Observed edges that were successfully preserved in the generated graphs.
    \item \textbf{Solid red lines}: Observed edges that were \textit{not} preserved in the generated graphs.
    \item \textbf{Dotted green lines}: Observed non-edge entries that were correctly preserved as non-edges.
    \item \textbf{Dotted red lines}: Observed non-edge entries that were \textit{not} preserved (i.e., incorrectly generated as edges).
\end{itemize}
As observed in the figure, graphs generated by our GGDiff methods exhibit a clear prevalence of green lines (indicating high preservation of observed entries and edges), whereas the unconstrained case shows a greater number of red lines, highlighting its inability to reliably reproduce the specified topological constraints.

\begin{table}[h]
\centering
\caption{Results for the incomplete graph generation experiment with observed edges.}
\label{tab:incomplete_graph_gen_edges}
\begin{tabular}{ccccc}
\toprule
\multirow{2}{*}{\textbf{Method}} & \multicolumn{2}{c}{\textbf{QM9}} & \multicolumn{2}{c}{\textbf{ZINC250k}} \\ \cmidrule(r){2-3} \cmidrule(r){4-5}
& Acc. (\%) & \% Unique & Acc. (\%) & \% Unique \\
\midrule
GGDiff-G & 79.73 & 86.53 & 95.98 & 100.00 \\
GGDiff-C & 29.18 & 97.56 & 19.73 & 99.90 \\
GGDiff-Z & 41.66 & 92.71 & 85.59 & 100.00 \\
Uncons. & 23.40 & 97.81 & 8.45 & 100.00 \\
\bottomrule
\end{tabular}
\end{table}

\section{Social impacts} \label{app:social_impacts}
The generation of graphs under constraints could lead to undesired consequences if not applied with care. 
For example, when doing graph completion.
In sensitive applications like healthcare or finance, these prediction inaccuracies can have serious repercussions, including misdiagnosis or financial losses. 
Moreover, misusing these models in social network analysis might inadvertently reinforce biases or invade privacy if not handled ethically. 
Therefore, it is crucial to apply GGDiff and any other graph inference algorithm using diffusion models with a thorough understanding of their limitations and to validate results rigorously to mitigate these risks.


\end{document}